\documentclass[a4paper]{article}

\PassOptionsToPackage{numbers}{natbib}
\usepackage{graphicx}

\usepackage[english]{babel}
\usepackage[utf8]{inputenc}

\usepackage[final]{nips_2017}

\usepackage[T1]{fontenc}    % use 8-bit T1 fonts
\usepackage{hyperref}       % hyperlinks
\usepackage{url}            % simple URL typesetting
\usepackage{booktabs}       % professional-quality tables
\usepackage{amsfonts}       % blackboard math symbols
\usepackage{nicefrac}       % compact symbols for 1/2, etc.
\usepackage{microtype}      % microtypography
\usepackage{color}
\usepackage{amsmath}
\usepackage[english]{babel}
\usepackage[utf8]{inputenc}
\usepackage{algorithm}
\usepackage[noend]{algpseudocode}
\usepackage{sidecap}

\usepackage{blindtext}

\newcommand\Algphase[1]{%
\vspace*{-.7\baselineskip}\Statex\hspace*{\dimexpr-\algorithmicindent-2pt\relax}\rule{\textwidth}{0.4pt}%
\Statex\hspace*{-\algorithmicindent}\textbf{#1}%
\vspace*{-.7\baselineskip}\Statex\hspace*{\dimexpr-\algorithmicindent-2pt\relax}\rule{\textwidth}{0.4pt}%
}

\newcommand\blfootnote[1]{%
  \begingroup
  \renewcommand\thefootnote{}\footnote{#1}%
  \addtocounter{footnote}{-1}%
  \endgroup
}

\title{Preserving Intermediate Objectives: One Simple Trick to Improve Learning for Hierarchical Models}

\author{
  Abhilasha Ravichander*\\
  Carnegie Mellon University\\
  \texttt{aravicha@cs.cmu.edu}\And Shruti Rijhwani*\\
    Carnegie Mellon University\\
  \texttt{srijhwan@cs.cmu.edu} \And Rajat Kulshreshtha*\\
    Carnegie Mellon University\\
  \texttt{rkulshre@cs.cmu.edu} \And Chirag Nagpal\\
    Carnegie Mellon University\\
  \texttt{chiragn@cs.cmu.edu} \And Tadas Baltru\v{s}aitis\\ 
    Carnegie Mellon University\\
  \texttt{tb346@cam.ac.uk} \And Louis-Philippe Morency\\
    Carnegie Mellon University\\
  \texttt{lmorency@andrew.cmu.edu}
}

\begin{document}
\blfootnote{*Indicated authors contributed equally}
\maketitle

\begin{abstract}
Hierarchical models are utilized in a wide variety of problems which are characterized by task hierarchies, where predictions on smaller subtasks are useful for trying to predict a final task. Typically, neural networks are first trained for the subtasks, and the predictions of these networks are subsequently used as additional features when training a model and doing inference for a final task.  In this work, we focus on improving learning for such hierarchical models and demonstrate our method on the task of speaker trait prediction. Speaker trait prediction aims to computationally identify which personality traits a speaker might be perceived to have, and has been of great interest to both the Artificial Intelligence and Social Science communities. Persuasiveness prediction in particular has been of interest, as persuasive speakers have a large amount of influence on our thoughts, opinions and beliefs. In this work, we examine how leveraging the relationship between related speaker traits in a hierarchical structure can help improve our ability to predict how persuasive a speaker is. We present a novel algorithm that allows us to backpropagate through this hierarchy. This hierarchical model achieves a 25\% relative error reduction in classification accuracy over current state-of-the art methods on the publicly available POM dataset.
\end{abstract}
 
\section{Introduction}

With the advent of social media and video-sharing websites like YouTube, multimedia has transformed from a carefully curated source to a constant background steam of information that has become ubiquitous in our lives.  Thus, understanding what makes certain online speakers influential to a wider audience is of considerable interest\cite{Schuller2012, metze2011review, rapp2011aristotle, matthews2003personality}. In this work, we explore how leveraging the relationship between related speaker traits such as passion(if a speaker is perceived to convey intense emotion in their content), or credibility(if the speaker is perceived as being worthy of trust) in a task hierarchy can help improve classification performance for persuasiveness. The motivation behind choosing passion and credibility as our initial subtasks stems from Aristotle's philosophy that the key to persuasive communication lies in \textit{ethos} (credibility), \textit{pathos} (passion) and \textit{logos} (logical cogency)~\cite{aristotle}. Further, significant social psychology research has analyzed and established the inter-dependencies between the perceived passion, credibility and persuasiveness or a speaker~\cite{garsten2009saving, burgoon1990nonverbal, pornpitakpan2004persuasiveness}.

While there has been considerable prior work on personality traits in the traditional context of psychology~\cite{barrick1991big}, research towards computationally identifying and predicting these traits is limited. To that end, recent progress in the computer vision, speech, and natural language processing communities has enabled us to devise methods to  computationally represent these cues from the visual, acoustic, and linguistic modalities. We attempt to combine these cues in order to predict speaker persuasiveness. Unlike prior work\cite{Park2014ComputationalAO,nojavanasghari2016deep}, we extend our model to \emph{ternary} persuasiveness classification (positive, negative, neutral) which is a much more challenging task due to the neutral class, as well as much more representative of the real world, where most videos are not just strongly persuasive and not persuasive at all. 

Task hierarchies are typically modeled using the stacking ensembling technique, by training classifiers for subtasks as using their predictions as features during training and inference for a final task. However, the models for the subtasks are static, and do not participate in the training for the final task. In this work, we propose a new algorithm which allows us to backpropagate through the stack while simultaneously preserving objectives at intermediate layers of a neural network, thus enabling us to get new state-of-the-art results for ternary persuasiveness prediction. The model serves to preserve (and potentially improve) the prediction quality of intermediate speaker traits while improving the final top-level classifier. Such a model also lends itself easily to semi-supervised learning settings where little data for training the intermediate objectives is available. Although in this work, we  utilize this model for speaker trait prediction specifically, it is broadly applicable to other tasks which have hierarchical structure where the final prediction depends on an intermediate prediction, such as dependency parsing with an intermediate objective for POS-tagging or facial expression recognition with intermediate objectives for Action Unit classification.

 %\rk{where the final prediction depends on an intermediate prediction : do we want to define hierarchical in this context? }

This paper is organized as follows -- in Section~\ref{related}, we examine related work in the field of computationally identifying speaker traits. In Section~\ref{proposed} we describe our proposed models and in Section~\ref{experiments} we describe the details of our experimental setup. Section~\ref{results} presents our results and discuss interesting insights and we conclude in Section~\ref{conclusion}.

\section{Related Work}
\label{related}
Identification and analysis of high-level speaker traits has a rich basis in social psychology. Passion and credibility, in particular, are noted to be important characteristics for eloquent speech \cite{aristotle}. Garsten et al. \cite{garsten2009saving} discuss the importance of passion in rhetoric, and several other studies have analyzed the relationship between credibility and persuasion \cite{burgoon1990nonverbal, pornpitakpan2004persuasiveness}. Ekman et al. \cite{ekman} explore the role face, body and speech play in judgments of personality and effect, and Kleinke et al. \cite{gaze} studied gaze and how it relates to making conciliatory and demanding requests. 

Several speech and language traits have also been analyzed using computational methods. Argamon et al. \cite{Argamon05lexicalpredictors} study lexical predictors of personality from text samples.
Further, Biel et al.~\cite{facetube} try to characterize video loggers using facial expressions and the Big-Five traits~\cite{john1999big}. Niculae et al. \cite{niculae2015linguistic} and Danescu et al. \cite{danescu2013computational} use textual language to predict betrayal and politeness, respectively. %\rk{considering that the following section is more multimodal and this section is unimodal, remove this line altogether: Using multimodal representations, Abouelenien et al.~\cite{deception} model deceit of a speaker. The study observes that a multimodal approach yields better performance than any single modality. Aran et al. \cite{meetings} use audiovisual recordings to infer speaker personality traits.}

There has also been considerable research in leveraging multiple modalities in order to identify high-level speaker traits, like passion, credibility and persuasion. The Persuasive Opinion Multimedia (POM) dataset was introduced by Park et al. \cite{park2014computational}, where they also discuss how descriptors from each modality can be used for predicting persuasiveness. Their experiments (using an SVM) show that multimodal techniques fare better than unimodal models. To identify if a speaker is perceived as passionate and/or credible, Chatterjee et al. \cite{Chatterjee2015CombiningTP} propose an ensemble classification approach which combines two models -- one that assumes inter-modality conditional independence, and one that explicitly represents the correlation between the different modalities in a lower dimensional subspace. They also use the Doc2Vec \cite{mikolov} representation to capture the semantic content of the text modality. Mohammadi et al. \cite{mohammadi_who_2013} predict whether a speaker is perceived as persuasive by taking into account three modalities. They also examine the effect of each modality and how persuasion ties into personality prediction. In \cite{Park:2016:MAP:2997043.2897739}, Park et al. observe that passion and credibility can help improve the results of persuasiveness prediction. Using a novel deep fusion technique, Nojavanasghari et al. \cite{nojavanasghari2016deep} classify persuasiveness of the speaker. They experiment with both early fusion and late fusion for prediction, and find that the latter performs better. Siddiquie et al. \cite{politics} discuss identification of \emph{politically persuasive} videos using the Rallying a Crowd dataset \cite{ROC}.

Maragos et al. \cite{maragos2008multimodal} discuss various methods to integrate information from different modalities. Early fusion is defined as concatenating all unimodal features into a single aggregate multimodal descriptor, whereas approaches for late fusion learn semantic concepts from unimodal features, which are subsequently used for learning the overall objective. Chatterjee et al. \cite{Chatterjee2015CombiningTP} state that while early fusion techniques combine cues from multiple modalities, they do not explicitly model the inter-modality correlations. Deep learning in multiple modalities has been used in several applications, like speech recognition \cite{ngiam2011multimodal}, retrieval \cite{srivastava2012multimodal}, and predicting personality traits like persuasion \cite{nojavanasghari2016deep}.

We also propose a novel neural architecture which leverages information across multiple modalities and intermediate tasks to achieve better classification performance on the final task, while simultaneously also improving it's performance on the intermediate tasks. While this work is inspired by work in multitask learning\cite{Caruana:1997:ML:262868.262872,abu1990learning,collobert2011natural}, it perhaps is most closely connected to the stacking ensembling technique where a learning algorithm utilizes the predictions of many other learning algorithms as features. The principle difference is that in our approach, the models for the subtasks are not static and learning happens continuously as we allow backpropagation through the stack. While we experiment with the final task of persuasiveness prediction, the architecture we present is generalized and can be used for any similar task setting.
\section{Method}
\label{proposed}
%We approach the problem of predicting speaker traits by modeling the relationships between the personality traits as a hierarchy and utilizing predictions of related traits while predicting the final trait (\textbf{Stacking}).

%However, traditional stacking simply uses classifier predictions as features for the final classifiers. We cannot backpropagate through the stack, and thus do not benefit in any way from information about the final task. In this paper, we propose a novel algorithm for stacking which preserves objectives at intermediate layers of the neural network, thus allowing us to backpropagate all the way through the stack. 

\subsection{Hierarchical Model (Stacking)}

\begin{SCfigure}
\centering
\includegraphics[width=0.5\textwidth]{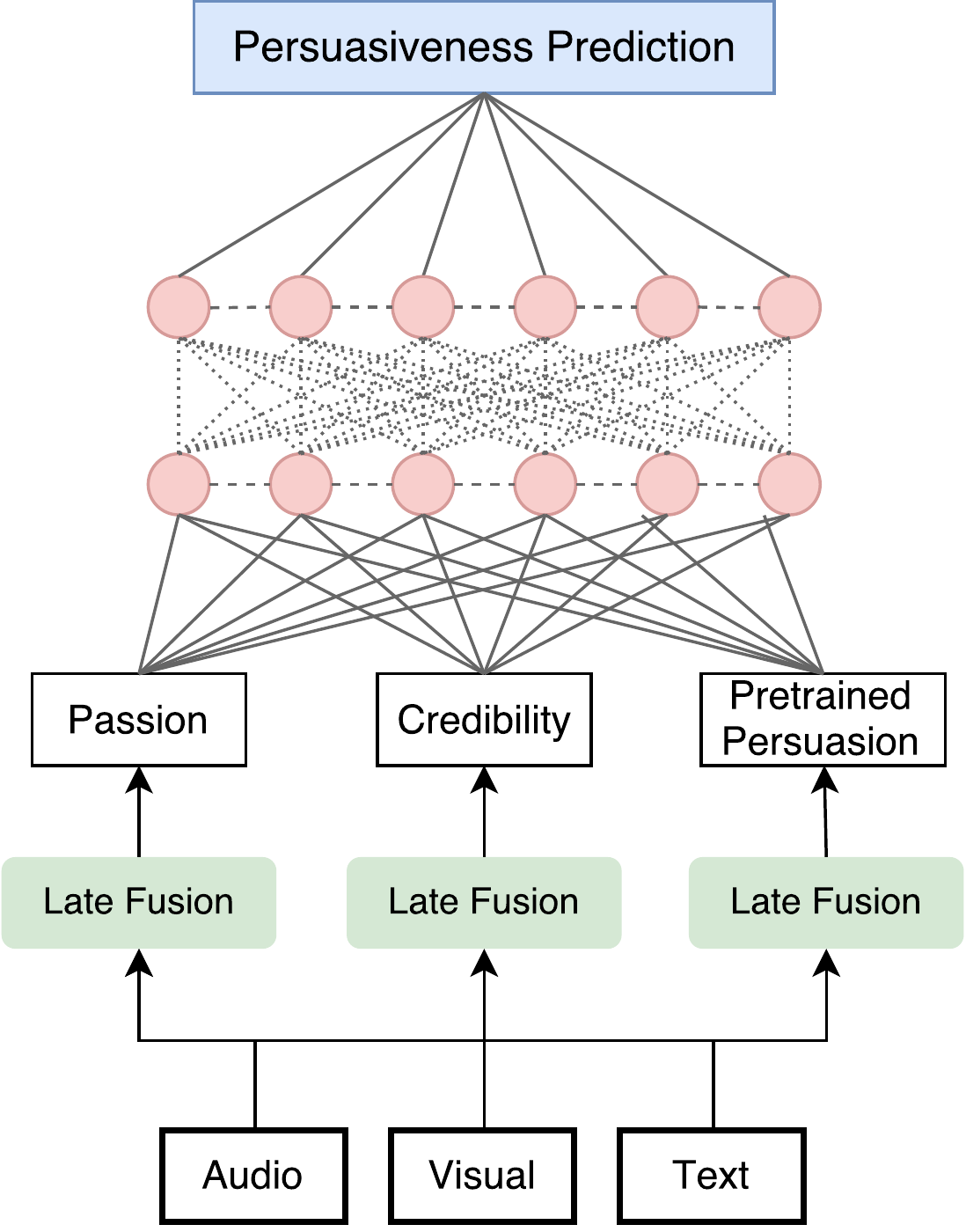}
\caption{Neural Hierarchical Model for persuasiveness prediction. The model consists of a passion network, a credibility network and an intermediate network that reduces the modalities to a lower dimensional representation with a persuasion objective. The last layer of the intermediate persuasion network is popped out and the previous layer's representation is used. The final persuasion network composes these models. The passion, credibility and intermediate persuasion networks are pretrained for weights that reduce their respective losses. For the hierarchical model with intermediate-objective preserving constraints, we only preserve objectives for the passion and credibility networks. For a simple stacking model, we set the acceptable error rate to $\infty$ and for end-to-end training we set the error rate to $1.0$}
\label{fig:hier}
\end{SCfigure}

We model the relationship between speaker traits in the form of a graph with dependencies. In the hierarchical model, we consider persuasiveness to be the final objective of our prediction. The hierarchical model first includes intermediate classifiers for passion and credibility and utilizes their predictions as inputs to the final classifier. The architecture is shown in Figure~\ref{fig:hier}. The training algorithm is as described. This utilizes the typical stacking ensembling technique where classifiers are first trained to predict passion and credibility, and their predictions are subsequently utilized as additional features during training and inference for persuasiveness prediction.%providing new state-of-the-art results for ternary classification on the POM dataset.

\begin{algorithm}
\caption{Training Algorithm for Stacking}
\begin{algorithmic}[1]
\Procedure{Stacking}{$D, L, p, c, pi, pe$}\Comment{Input Dataset $D$ with training labels and their target trait scores(X,Y), learning rate $L$, passion network $P$, credibility network $C$, intermediate persuasion network $Pi$ and final persuasion network $Pe$.}
\Algphase{Phase 1 - Training networks for subtasks}
\State P: $L_{passion}\gets \sum_{i=1}^{n}-log(P(y_{passion,i}))$ \Comment{Train network for subtask of passion prediction (\emph{P}).}
\State C: $L_{credibility}\gets \sum_{i=1}^{n}-log(P(y_{credibility,i}))$ \Comment{Train network for subtask of credibility prediction (\emph{C}).}
\Algphase{Phase 2 - Pretrain intermediate network for final task}
\State Pi: $L_{persuasion}\gets \sum_{i=1}^{n}-log(P(y_{persuasion,i}))$ \Comment{Train network for persuasion prediction (\emph{pi}).}
\State Pi: Pop last layer of $\emph{Pi}$
\Algphase{Phase 3 - Training final network}
\State Compose persuasion network \emph{Pe} with passion (\emph{P}) and credibility (\emph{C}) networks and intermediate persuasion network (\emph{Pi}), followed by a fully connected network of 2 or more layers as shown in Figure 2.
\State P, C, Pi, Pe: $L_{persuasion}\gets \sum_{i=1}^{n}-log(P(y_{persuasion,i}))$ \Comment{Train persuasion network $Pe$ with modality features as well as predictions from $p$ and $c$ as input}
\EndProcedure
\end{algorithmic}
\end{algorithm}

\subsection{Intermediate-Objective preserving Hierarchical Model (HIO)}

In this work, we present a novel ensembling model as an alternative to stacking, which retains the advantages of stacking while also being end-to-end trainable (Figure~\ref{fig:hier}). Stacking is a popular approach for tasks with dependencies, where a to-level classifier has access to not only the input features from the video but also the predictions of intermediate classifiers as input before making it's final prediction. However, the intermediate networks are static, cannot be backpropagated through and thus do not benefit in any way from information of the final task.

\begin{algorithm}
\caption{Training Algorithm for HIO}
\begin{algorithmic}[1]
\Procedure{HIO}{$D, L, p, c, pi, pe, \epsilon$}\Comment{Input Dataset $D$ with training labels and their target trait scores (X,Y), Acceptable Error Rate ($\epsilon$), learning rate $L$, passion network $P$, credibility network $C$, intermediate persuasion network $Pi$ and final persuasion network $Pe$.}
\Algphase{Phase 1 - Training networks for subtasks}
\State P: $L_{passion}\gets \sum_{i=1}^{n}-log(P(y_{passion,i}))$ \Comment{Train network for subtask of passion prediction (\emph{P}).}
\State C: $L_{credibility}\gets \sum_{i=1}^{n}-log(P(y_{credibility,i}))$ \Comment{Train network for subtask of credibility prediction (\emph{C}).}
\Algphase{Phase 2 - Pretrain intermediate network for final task}
\State Pi: $L_{persuasion}\gets \sum_{i=1}^{n}-log(P(y_{persuasion,i}))$ \Comment{Train network for persuasion prediction (\emph{pi}).}
\State Pi: Pop last layer of $\emph{Pi}$
\Algphase{Phase 3 - Training final network}
\State Compose persuasion network \emph{Pe} with passion (\emph{P}) and credibility (\emph{C}) networks and intermediate persuasion network (\emph{Pi}), followed by a fully connected network of 2 or more layers as shown in Figure 2.
\State P, C, Pi, Pe: $L_{persuasion}\gets \sum_{i=1}^{n}-log(P(y_{persuasion,i}))$ \Comment{Train end-to-end through \emph{Pe} to predict persuasion}
\Procedure{Backpropagation}{}
\State $\partial w_{t, i,j} \gets l \cdot \frac{dl}{dw}$
\State $w_{t+1,i,j} = w_{t, i,j} + \partial w_{t,i,j}$
\State $L^{'}_{passion} = \sum_{i=1}^{n}-log(P(y_{i})) $ \Comment{Do an imaginary forward pass and check loss for passion on validation data}
\State $L^{'}_{credibility} = \sum_{i=1}^{n}-log(P(y_{i})) $ \Comment{Do an imaginary forward pass and check loss for credibility on validation data}
\For{$w_{i,j} \in \{W_{P}, W_{C}\}$}\Comment{$W_{P}$ is the weight matrix of the passion network, and $W_{C}$ is the weight matrix for the credibility network}
\If{$L^{'}_{passion} \leq \epsilon \cdot L_{passion}$}
\State no change (keep the new weights)
\Else
\State $w_{t+1,i,j} \gets W_{t,i,j}$ \Comment{Revert weights to before weight update} 
\EndIf
\EndFor

\State $t \gets t+1$
\EndProcedure

\EndProcedure
\end{algorithmic}
\end{algorithm}

%of a neural network while optimizing for a final task.
Our approach seeks to allow gradient flow even through these intermediate networks, while still preserving objectives at intermediate layers.  This model is attractive for a variety of tasks which have hierarchical structure. The model consists of two key stages, an \emph{imaginary forward pass} through the validation data to estimate distance from the intermediate objectives and a \emph{gated weight update} which prevents gradient-flow for constraint-violating updates. The model introduces a hyperparameter in the form of the \emph{Acceptable Error Rate} which is the degree of error we are willing to accept from our constraint. The intuition behind this hyperparameter is that it represents the extent to which we try to preserve objectives at the intermediate layers of the neural network. In the strictest case we will set this to 1 and thus only ever update when the new weights decrease our validation loss. 

The size of our dataset can also affect the value we choose for this hyperparameter. For small datasets, spurious weight updates can have a large effect on performance. We set the \emph{Acceptable Error Rate} to be 1.0, forcing strict monotonicity in the validation accuracies for the individual networks. For larger datasets where we can afford to occasionally make mistakes, the \emph{Acceptable Error Rate} could be treated as a hyperparameter representing a tradeoff between final-objective optimization and intermediate-objective preservation. Stacking can also be viewed as a special case of this algorithm, with the Acceptable Error Rate being set to $\infty$. 

This model is also attractive from the standpoint of semi-supervised learning, wherein intermediate objectives could have significantly lesser data than the final objective. For example, in the case of Action Units for Facial Expression Recognition, Action Units are considerably harder to annotate (and in fact require training to identify), as opposed to facial expressions. Thus, with sparsely trained intermediate networks, the final objective could even be used to improve the performance of intermediate networks since our model allows backpropagation through intermediate networks.

We will now follow with an explanation of the intuition behind each of the steps in the training algorithm:\\
Training Algorithm
Input: Dataset $D$ with training labels and their target trait scores, learning rate \emph{L} , acceptable error rate $\epsilon$, passion network \emph{P}, credibility network \emph{C}, intermediate persuasion network \emph{Pi}, and final persuasion network \emph{Pe}.\\
Algorithm:
\begin{enumerate}
\item Train passion network \emph{P}:
$$L_{passion} = \sum_{i=1}^{n}-log(P(y_{i}))$$ where $P(y_{i})$ is the probability of the true class for a given sample.\\
This step trains a network to classify a speaker into one of three classes (very passionate, mildly passionate, not passionate), based on how passionate they are perceived to be.
\item Train credibility network \emph{C}:
$$L_{credibility} = \sum_{i=1}^{n}-log(P(y_{i}))$$ where $P(y_{i})$ is the probability of the true class for a given sample.\\
This step trains a network to classify a speaker into one of three classes (very credible, mildly credible, not credible), based on how credible they are perceived to be.

\item Train intermediate persuasion network \emph{Pi}  with modality features from video and persuasion objective. Pop last layer after training.

This step trains a persuasion network to classify speakers based on their persuasiveness. The last layer is removed. Thus at the end of this step, we have essentially performed some pretraining and brought the weights of this smaller network to a space that is useful to predict persuasion.\\
At the end of this stage stage, we have a pretrained passion network \emph{P}, credibility network \emph{C} and intermediate persuasion network \emph{Pi}.
\item Compose persuasion network \emph{Pe} with passion (\emph{P}) and credibility (\emph{C}) networks and intermediate persuasion network (\emph{Pi}), followed by a fully connected network of 2 or more layers as shown in Figure 2.
\item Train end-to-end through \emph{Pe} to predict persuasion
$$L_{persuasion} = \sum_{i=1}^{n}-log(P(y_{i}))$$
\item Backpropagation Step through all layers of the network \emph{Pe}
$$\partial w_{i,j} = l \cdot \frac{dl}{dw}$$
$$w_{i,j} = w_{i,j} + \partial w_{i,j}$$
We backpropagate completely through the overall persuasion network. This includes the fully-connected layers, the passion network, the credibility network and the intermediate persuasion network
\item Imaginary Forward Pass with new weights for networks \emph{P} and \emph{C}
$$L^{'}_{passion} = \sum_{i=1}^{n}-log(P(y_{i})) $$\\

Similarly for credibility. We do an imaginary forward pass with the validation data for the network of each of our subtasks and compute the loss.
\item Gated Update with
Acceptable Error Rate ($\epsilon$) for networks \emph{P} and \emph{C}\\

if $L^{'}_{passion} \leq \epsilon \cdot L_{passion}$: \\
no change (keep the new weights)\\

else: \\
$W_{t+1} = W_{t}$ (revert to weights from the previous step)\\

Similarly for credibility.

We examine if the loss computed in the imaginary forward pass for each of the smaller subtasks with the new weights violates our constraint. If these weights do violate the constraint, we revert back to the previous weights. In this way we backpropagate through the network, only allowing weight updates that we find acceptable. This allows us to completely backpropagate through the stack while simultaneously preserving intermediate objectives. We can choose to compute this loss (and consequently the constraint) on either the training data or the validation data. In practice we find that using the validation data works best, thus we can also draw connections between our method and early stopping.
\end{enumerate}

Note that the imaginary forward pass and gated update constraint applies only for the models of intermediate tasks. The intermediate models are allowed to be updated when the gate condition is satisfied.

\section{Experiments}
\label{experiments}
In this work, we consider the problem of persuasiveness prediction for three classes of persuasiveness - very persuasive, not persuasive and neutral. We partition our data into ten \textit{speaker-independent folds} and perform cross-validation on them. We use cross-validation to account for variance across test-sets, by considering 90\% as train set and 10\% as test set (9 fold and 1 fold). We randomly pick one fold from the training set and use it as the validation set at each experiment. This is primarily used for early stopping, where we save the best model over training epochs as measured on the validation accuracy.

\subsection{Dataset}
The dataset we  use is the \textsc{Persuasive Opinion Multimedia} (POM) dataset \cite{Park2014ComputationalAO}. 
The corpus contains a thousand movie review videos gathered from \emph{ExpoTV.com}. Each video is annotated with its persuasiveness, along with confidence, credibility, dominance, humor, and passion, amongst other speaker traits, by three human annotators. The ratings are on a Likert scale~\cite{likert1932technique} of $1$ to $7$. The Pearson's correlation coefficient between the annotations of passion and persuasion and credibility and persuasion are $0.55$ and $0.73$ respectively.

In this work, we focus on predicting persuasiveness. We consider the average rating obtained for each trait as the trait score. For our task of persuasiveness classification, we consider videos with an average rating of less than $3$ as negative or greater than $5$ as positive, and a rating between $3$ and $5$ inclusive as neutral on each of the traits. 

We use the visual and acoustic feature descriptors available in POM dataset~\cite{Park2014ComputationalAO}. We also extract features from OpenSmile features for audio~\cite{Eyben2013}. Since the audio and video features are computed for short time units, we use the mean, standard deviation, minimum, maximum, and the max-min range to represent these in a definite-sized input for each data point. For text features, we use TF-IDF from the review transcripts. After extracting the features, we use the t-test feature selection method as described in \cite{nojavanasghari2016deep} to reduce the learning space to the relevant features. 

\subsection{Multimodal Baseline}
For our multimodal baseline, we use the deep late fusion model designed by Nojavanasghari et al.~\cite{nojavanasghari2016deep} since it has the best performance for persuasion prediction on the POM dataset. The model trains a multilayer perceptron (MLP) for each modality and then fuses the outputs of the modalities to form the input for another MLP that performs the final persuasion classification. However, the model was originally designed for binary persuasion classification (highly positive and highly negative). In order to adapt it for our problem of ternary classification, we train the modality-specific MLPs to predict three values and use these as a vector input to the fusion MLP~\footnote{We tested our reimplementation on binary classification and could match the results reported in~\cite{nojavanasghari2016deep}.}. The cross-validation accuracy is reported in Table~\ref{resultstable}. 

We use this baseline as the late fusion model for pretraining the intermediate trait networks in our hierarchical algorithms (Figure~\ref{fig:hier}).

\subsection{Models}
We implement two hierarchical models as described in section 3.1 and 3.2.
\subsubsection{Stacking}
The first is a simple stacking model. The modalities are combined using a deep fusion model as described in section 4.2. We then train a passion network consists of 5 layers, each with 5 hidden units an a \textsc{Relu} activation, and a credibility network that has the same architecture. We also pretrain a third network to predict persuasion with features from all three modalities, pop the last layer and utilize this representation as input to a  final neural network trained to predict persuasion. That is, we fuse the penultimate layer of this network with the output of the passion network, the credibility network. This forms the input of our persuasion model which consists of 3 layers with 5 hidden units each and a categorical cross-entropy loss. At test time, the prediction of the passion and credibility networks are used as features in addition to the features extracted from the different modalities.

\subsubsection{Hierarchical Intermediate Objective (HIO)}
The second model we implement is a hierarchical model that preserves objectives at intermediate layers of the neural network, while still allowing weight updates. In this case, these are the objectives for passion and credibility for networks trained to predict each of these traits. The passion network consists of 5 layers, each with 5 hidden units an a \textsc{Relu} activation. The credibility network has the same architecture. The modalities are combined using a deep fusion model as described in section 4.2. We then feed this representation as input to a neural network trained to predict persuasion. We fuse the penultimate layer of this network with the output of the passion network and the credibility network, and use this as the input representation for our final persuasion model which consists of 3 layers with 5 hidden units each and a categorical cross-entropy loss.
We perform early stopping on each of the networks, storing weights at every 10th epoch.

We run this model with an \emph{Acceptable Error Rate} of $\infty$ for a hierarchical model (unconstrained) and $1.0$ using the algorithm presented in section 3.3.
The results of this model are presented below.

\begin{center}
\begin{minipage}[t]{.5\linewidth}
\vspace{0pt}
\centering
\includegraphics[width=\linewidth]{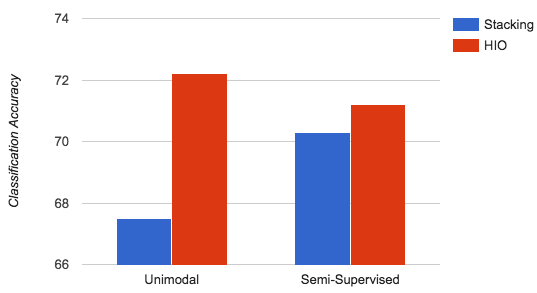}
\label{unisemi}
\end{minipage}
\begin{minipage}[t]{.5\linewidth}
\vspace{0pt}
\centering
\begin{tabular}{l c}
\hline
Model                                                                                                          & Accuracy \% \\ \hline
Baseline\cite{nojavanasghari2016deep}                                                                                         & 60.6     \\ 
Stacking                                                                                    & 69.53    \\ 
HIO  & \textbf{70.4}     \\ \hline

\end{tabular}
\label{resultstable}

\end{minipage}
\end{center}

In Figure \ref{unisemi}, we observe the classification performance Hierarchical Intermediate Objective (HIO) vs. Stacking for unimodal (with text) and semi-supervised learning. We observe that the HIO model gives considerable improvements in these settings. In Table \ref{resultstable}, we present the multimodal late fusion accuracy for all models on ternary persuasiveness classification.

\subsection{Semi-Supervised Learning}
As a final test of the potential of our hierarchical constrained model, we experiment with a semi-supervised learning setting, in order to see whether the system is robust to situations where the lower-level models (passion and credibility, in our case) are not pre-trained with a lot of data. 

Instead of using 8 training folds, we pre-train the passion and credibility models using the late fusion technique with just 2 training folds each. We then use these models in both our hierarchical models. The results are shown in Table~\ref{resultstable} and we notice that this setting with fewer training samples performs similarly to our experiments with the whole training set.

\section{Results and Discussion}
\label{results}

The results from our experiments are detailed in Table~\ref{resultstable}. As stated earlier, the research question we aim to answer is whether the relationship between different speaker traits can be leveraged to improve the prediction accuracy of persuasiveness.

During our experiments we observed that prediction on the POM dataset is potentially unreliable due to the small size of the test set, and performance varies highly depending on the choice of the test set. 

To avoid reporting artificially high results, we choose to perform cross validation by considering each of our ten speaker-independent folds so the test set is unbiased. The cross-validation causes the accuracy to stabilize across random restarts of training. 

In addition, we treat persuasiveness prediction as a ternary classification problem instead of a binary one, making the task more challenging, but also more realistic as speaker traits are not always highly polar. We experimented with undersampling and oversampling to account for the class imbalance caused by a large neutral class, but due to the small size of the POM dataset we found that this leads to slightly worse performance. This is likely because using t-test leads selecting features that are reasonably discriminative, rendering sampling is unnecessary.

We observe that in the vanilla multitask model and hierarchical multitask model, reducing all the speaker traits to sharing a single parameter space over the modalities does not benefit the prediction of persuasiveness. Although it is significantly highly in accuracy than randomly selecting a score (33\% chance), it is over 10\% lower in accuracy than the multimodal baseline. Further, we notice that adding passion and credibility features to the persuasiveness prediction model in the hierarchical multitask improves the cross-validation accuracy. Intuitively, these observations could mean that, even though the perception of a trait might influence another trait, the traits depend differently on each modality and cannot be predicted well with shared parameters. For example, an animated person (visual) with a bright tone (acoustic) may be perceived as passionate, but is not necessarily credible.

We observe that the hierarchical model structure lends itself to strong performance in predicting speaker persuasiveness, and that passion and credibility predictions help to significantly improve this performance. The model with intermediate-objective preserving constraints maintains strong monotonicity in performance, giving the same or better results compared to simple hierarchical models. We see an enormous improvement in accuracy over the baseline model, which signifies that predicting persuasion is benefited by using the inputs from correlated speaker traits. Further, we notice that adding the intermediate-objective-preserving constraint improves the accuracy slightly. In contrast to the multitask scenario, the models for each trait are pre-trained individually and then merged to form a single model. Although all weights are updated through backpropagation after joining the models, the parameters are not shared which allows the model to learn different dependencies on modalities for each trait. 

We also notice that the model trained on just the text features performs comparably to the multimodal models in both the simple MLP case (compared to late fusion) and the hierarchical case. We hypothesize that the text features themselves are extremely discriminative of the small number of examples in the POM dataset and hence do not gain much from other modalities. However, we would like to explore the benefits of each modality in future work.

In addition, a preliminary exploration of semi-supervised learning where the model is only shown very limited passion and credibility data, yields promising results. We note that this would be particularly useful in scenarios where annotations for the intermediate properties are scarce, but the final objective is easily annotated. For example, passion and credibility are abstract concepts, but persuasiveness of a movie review can be determined by the sentiment of the review (positive or negative) and whether the reviewer convinced the annotator to watch (or not watch) the movie~\cite{Park2014ComputationalAO}.

% Variance in POM
% Ternary

%Constraint model
%Semi-supervised
%prior work
\section{Conclusion and Future Direction}
\label{conclusion}
For the current work, given the basis in social science as well as the high correlation in data between the traits of passion and credibility with persuasion,  we focused on the relation between these three traits. In the future, we would like to see whether we can extend this model to model other high level speaker traits, or possibly even other speaker attributes (like age, gender). This could possibly enable us to test our models on more datasets, as well as real world data like political debates or advertisements. %\rk{Being able to perform such analysis is the ultimate end goal for such techniques, and it will enable many applications in the fields of advertising and public policy}.

In addition, we believe that our model which preservers intermediate objectives will generalize well to other tasks that have a hierarchical structure such as facial expression recognition, as well as in semi-supervised settings. We would like to experiment with more data that exhibits these characteristics as well as further explore our technique's application in semi-supervised learning.
%Perhaps most significantly, we believe that the proposed intermediate-objective preserving model %can also be generalized to other tasks where we have some prior knowledge about the relationships between some intermediate labels and final targets. %By explicitly modeling this relationship, we can improve the performance on intermediate tasks in situations where there is not enough data available compared to the final task. 

\bibliographystyle{plain}
\bibliography{sample}

\begin{thebibliography}{10}

\bibitem{abu1990learning}
Yaser~S Abu-Mostafa.
\newblock Learning from hints in neural networks.
\newblock {\em Journal of complexity}, 6(2):192--198, 1990.

\bibitem{Argamon05lexicalpredictors}
Shlomo Argamon, Sushant Dhawle, Moshe Koppel, and James~W. Pennebaker.
\newblock Lexical predictors of personality type.
\newblock In {\em Proceedings of the Joint Annual Meeting of the Interface and
  the Classification Society of North America}, 2005.

\bibitem{barrick1991big}
Murray~R Barrick and Michael~K Mount.
\newblock The big five personality dimensions and job performance: a
  meta-analysis.
\newblock {\em Personnel psychology}, 44(1):1--26, 1991.

\bibitem{facetube}
Joan-Isaac Biel, Lucia Teijeiro-Mosquera, and Daniel Gatica-Perez.
\newblock Facetube: predicting personality from facial expressions of emotion
  in online conversational video.
\newblock In Louis-Philippe Morency, Dan Bohus, Hamid~K. Aghajan, Justine
  Cassell, Anton Nijholt, and Julien Epps, editors, {\em ICMI}, pages 53--56.
  ACM, 2012.

\bibitem{burgoon1990nonverbal}
Judee~K Burgoon, Thomas Birk, and Michael Pfau.
\newblock Nonverbal behaviors, persuasion, and credibility.
\newblock {\em Human communication research}, 17(1):140--169, 1990.

\bibitem{Caruana:1997:ML:262868.262872}
Rich Caruana.
\newblock Multitask learning.
\newblock {\em Mach. Learn.}, 28(1):41--75, July 1997.

\bibitem{Chatterjee2015CombiningTP}
Moitreya Chatterjee, Sunghyun Park, Louis-Philippe Morency, and Stefan Scherer.
\newblock Combining two perspectives on classifying multimodal data for
  recognizing speaker traits.
\newblock In {\em ICMI}, 2015.

\bibitem{ROC}
D.~Chisholm, B.~Siddiquie, A.~Divakaran, and E.~Shriberg.
\newblock Audio-based affect detection in web videos.
\newblock In {\em 2015 IEEE International Conference on Multimedia and Expo
  (ICME)}, pages 1--6, June 2015.

\bibitem{collobert2011natural}
Ronan Collobert, Jason Weston, L{\'e}on Bottou, Michael Karlen, Koray
  Kavukcuoglu, and Pavel Kuksa.
\newblock Natural language processing (almost) from scratch.
\newblock {\em Journal of Machine Learning Research}, 12(Aug):2493--2537, 2011.

\bibitem{danescu2013computational}
Cristian Danescu-Niculescu-Mizil, Moritz Sudhof, Dan Jurafsky, Jure Leskovec,
  and Christopher Potts.
\newblock A computational approach to politeness with application to social
  factors.
\newblock {\em arXiv preprint arXiv:1306.6078}, 2013.

\bibitem{ekman}
Paul Ekman, Wallace Friesen, Maureen O'Sullivan, and Klaus~R Scherer.
\newblock Relative importance of face, body, and speech in judgments of
  personality and affect.
\newblock 2014.

\bibitem{Eyben2013}
Florian Eyben, Felix Weninger, Florian Gross, and Bj\"{o}rn Schuller.
\newblock Recent developments in opensmile, the munich open-source multimedia
  feature extractor.
\newblock In {\em Proceedings of the 21st ACM International Conference on
  Multimedia}, MM '13, pages 835--838, New York, NY, USA, 2013. ACM.

\bibitem{garsten2009saving}
Bryan Garsten.
\newblock {\em Saving persuasion: A defense of rhetoric and judgment}.
\newblock Harvard University Press, 2009.

\bibitem{john1999big}
Oliver~P John and Sanjay Srivastava.
\newblock The big five trait taxonomy: History, measurement, and theoretical
  perspectives.
\newblock {\em Handbook of personality: Theory and research}, 2(1999):102--138,
  1999.

\bibitem{aristotle}
G.~A. Kennedy.
\newblock On rhetoric: A theory of civic discourse.
\newblock 1991.

\bibitem{gaze}
Chris~L Kleinke and David~A Singer.
\newblock Personality and social psychology bulletin.
\newblock 1979.

\bibitem{likert1932technique}
Rensis Likert.
\newblock A technique for the measurement of attitudes.
\newblock {\em Archives of psychology}, 1932.

\bibitem{maragos2008multimodal}
Petros Maragos, Alex Potamianos, and Patrick Gros.
\newblock {\em Multimodal processing and interaction: audio, video, text},
  volume~33.
\newblock Springer Science \& Business Media, 2008.

\bibitem{matthews2003personality}
Gerald Matthews, Ian~J Deary, and Martha~C Whiteman.
\newblock {\em Personality traits}.
\newblock Cambridge University Press, 2003.

\bibitem{metze2011review}
Florian Metze, Alan Black, and Tim Polzehl.
\newblock A review of personality in voice-based man machine interaction.
\newblock {\em Human-Computer Interaction. Interaction Techniques and
  Environments}, pages 358--367, 2011.

\bibitem{mikolov}
Tomas Mikolov, Ilya Sutskever, Kai Chen, Greg~S Corrado, and Jeff Dean.
\newblock Distributed representations of words and phrases and their
  compositionality.
\newblock In {\em NIPS}, 2013.

\bibitem{mohammadi_who_2013}
Gelareh Mohammadi, Sunghyun Park, Kenji Sagae, Alessandro Vinciarelli, and
  Louis-Philippe Morency.
\newblock Who {Is} {Persuasive}? {The} {Role} of {Perceived} {Personality} and
  {Communication} {Modality} in {Social} {Multimedia}.
\newblock In {\em Proceedings of the 15th {ACM} on {International} conference
  on multimodal interaction}, pages 19--26, New York, NY, December 2013. ACM
  Press.

\bibitem{ngiam2011multimodal}
Jiquan Ngiam, Aditya Khosla, Mingyu Kim, Juhan Nam, Honglak Lee, and Andrew~Y
  Ng.
\newblock Multimodal deep learning.
\newblock In {\em Proceedings of the 28th international conference on machine
  learning (ICML-11)}, pages 689--696, 2011.

\bibitem{niculae2015linguistic}
Vlad Niculae, Srijan Kumar, Jordan Boyd-Graber, and Cristian
  Danescu-Niculescu-Mizil.
\newblock Linguistic harbingers of betrayal: A case study on an online strategy
  game.
\newblock {\em arXiv preprint arXiv:1506.04744}, 2015.

\bibitem{nojavanasghari2016deep}
Behnaz Nojavanasghari, Deepak Gopinath, Jayanth Koushik, Tadas
  Baltru{\v{s}}aitis, and Louis-Philippe Morency.
\newblock Deep multimodal fusion for persuasiveness prediction.
\newblock In {\em Proceedings of the 18th ACM International Conference on
  Multimodal Interaction}, pages 284--288. ACM, 2016.

\bibitem{Park2014ComputationalAO}
Sunghyun Park, Han~Suk Shim, Moitreya Chatterjee, Kenji Sagae, and
  Louis-Philippe Morency.
\newblock Computational analysis of persuasiveness in social multimedia: A
  novel dataset and multimodal prediction approach.
\newblock In {\em ICMI}, 2014.

\bibitem{park2014computational}
Sunghyun Park, Han~Suk Shim, Moitreya Chatterjee, Kenji Sagae, and
  Louis-Philippe Morency.
\newblock Computational analysis of persuasiveness in social multimedia: A
  novel dataset and multimodal prediction approach.
\newblock In {\em Proceedings of the 16th International Conference on
  Multimodal Interaction}, pages 50--57. ACM, 2014.

\bibitem{Park:2016:MAP:2997043.2897739}
Sunghyun Park, Han~Suk Shim, Moitreya Chatterjee, Kenji Sagae, and
  Louis-Philippe Morency.
\newblock Multimodal analysis and prediction of persuasiveness in online social
  multimedia.
\newblock {\em ACM Trans. Interact. Intell. Syst.}, 6(3):25:1--25:25, October
  2016.

\bibitem{pornpitakpan2004persuasiveness}
Chanthika Pornpitakpan.
\newblock The persuasiveness of source credibility: A critical review of five
  decades' evidence.
\newblock {\em Journal of Applied Social Psychology}, 34(2):243--281, 2004.

\bibitem{rapp2011aristotle}
Christof Rapp.
\newblock Aristotle's rhetoric.
\newblock {\em Stanford Encyclopedia of Philosophy}, 2011.

\bibitem{Schuller2012}
B~Schuller, S~Steidl, A~Batliner, E~Neoth, A~Vinciarelli, and F~Burkhardt.
\newblock The interspeech 2012 speaker trait challenge.
\newblock In {\em Proceedings of Interspeech}, 2012.

\bibitem{politics}
Behjat Siddiquie, Dave Chisholm, and Ajay Divakaran.
\newblock Exploiting multimodal affect and semantics to identify politically
  persuasive web videos.
\newblock In {\em Proceedings of the 2015 ACM on International Conference on
  Multimodal Interaction}, ICMI '15, pages 203--210, New York, NY, USA, 2015.
  ACM.

\bibitem{srivastava2012multimodal}
Nitish Srivastava and Ruslan~R Salakhutdinov.
\newblock Multimodal learning with deep boltzmann machines.
\newblock In {\em Advances in neural information processing systems}, pages
  2222--2230, 2012.

\end{thebibliography}

\end{document}